%% file: main.tex
\documentclass{article}

\usepackage[numbers]{natbib}
\usepackage{graphicx}
\usepackage[preprint] {neurips_2023}

\usepackage[utf8]{inputenc} 
\usepackage[T1]{fontenc}    
\usepackage{hyperref}       
 \hypersetup{%
    unicode=true,%
    colorlinks=true,%
    citecolor=red,%
    linkcolor=blue,%
    urlcolor=red,
}
\usepackage{url}            
\usepackage{booktabs}       
\usepackage{amsmath,amsfonts}       
\usepackage{nicefrac}       
\usepackage{microtype}      
\usepackage{xcolor}         
\usepackage{comment}
\usepackage{placeins}
\usepackage{float}
\usepackage{bm}
\usepackage{subfig}
\usepackage{authblk}
\usepackage[compact]{titlesec}
\title{Mitigating Catastrophic Forgetting in\\ Long Short-Term Memory Networks}

\author{ 
\begin{minipage}[t, l]
{.30\linewidth}
\centering
\textbf{\centering Ketaki Joshi}

Department of Computer Science, Yale University
\texttt{ketaki.joshi@yale.edu}
\end{minipage}
\noindent
\begin{minipage} [t, l]
{.33\linewidth}
\centering 
\vspace{10pt}
\textbf{\centering Raghavendra Pradyumna 
Pothukuchi}

Department of Computer Science, Yale University
\texttt{raghav.pothukuchi@yale.edu}
\end{minipage}
\noindent
\begin{minipage} [t, l]
{.33\linewidth}
\centering 
\textbf{ \centering Andre Wibisono}

Department of Computer Science, Yale University
\texttt{andre.wibisono@yale.edu}
\end{minipage}

\centering
\begin{minipage} [t, l]
{.25\linewidth}
\centering
\textbf{\centering Abhishek 
Bhattacharjee}

Department of Computer Science, Yale University
\texttt{abhishek@cs.yale.edu}
\end{minipage}
}

\begin{document}
\maketitle
\input{0_abstract}

\input{1_Introduction}
\input{2_Background}
\input{3_RelatedWork}

\input{4_Hypothesis}

\input{5_Experiment_setup}

\input{6_Experimentation}
\input{Conclusion}

\newpage
\appendix
\input{7_SupplementaryMaterial}

\newpage
\bibliographystyle{unsrt}
\bibliography{ref}
\end{document}

%% file: 0_abstract.tex
\begin{abstract}
Continual learning on sequential data is critical for many machine learning (ML) deployments. Unfortunately, LSTM networks, which are commonly used to learn on sequential data, suffer from catastrophic forgetting and are limited in their ability to learn multiple tasks continually. We discover that catastrophic forgetting in LSTM networks can be overcome in two novel and readily-implementable ways---separating the LSTM memory either for each task or for each target label. Our approach eschews the need for explicit regularization, hypernetworks, and other complex methods. We quantify the benefits of our approach on recently-proposed LSTM networks for computer memory access prefetching, an important sequential learning problem in ML-based computer system optimization. Compared to state-of-the-art weight regularization methods to mitigate catastrophic forgetting, our approach is simple, effective, and enables faster learning.  We also show that our proposal enables the use of small, non-regularized LSTM networks for complex natural language processing in the offline learning scenario, which was previously considered difficult.  

\end{abstract}

%% file: 1_Introduction.tex
\section{Introduction}
\label{intro}

LSTM networks are commonly used to process sequential data in many application domains including biological and medical data processing, weather modeling, stock market analysis, language processing, dynamical systems optimization, and more~\cite{medicalLSTM,weatherLSTM,stockMarketLSTMPrediction,languageModellingWithLSTM, LearnMemAccPat, mlPrefetchervoyager}. In most real-world deployments, neural networks (including LSTM networks) must learn to process many sequential tasks on sequential data, where the order of the tasks and arrival times can be arbitrary. That is, they must learn continually---i.e., they should learn over new dynamically arriving tasks without jeopardizing their performance on tasks that had been previously learned for.

Unfortunately, however, LSTM networks (like other connectionist ML models) suffer from catastrophic forgetting, where the networks abruptly forget the ability to perform a task when they are trained on a new task~\cite{ehret2021continual,CFLSTM}. Consequently, LSTM and other networks are limited in their applicability to the many challenging real-world deployments.

Overcoming catastrophic forgetting is challenging. One might, for example, envision using the task identifier as an additional input to the neural network to mitigate catastrophic forgetting. Unfortunately, however, this approach is ineffective, particularly for LSTM networks, as it does not prevent the hidden weights from being completely overwritten across tasks~\cite{ehret2021continual}. In response, several recent studies have explored catastrophic forgetting in general feedforward networks (e.g.,~\cite{iCArL,ewc,PackNet}) and recurrent neural networks (RNNs) (e.g., LSTM networks~\cite{ehret2021continual, CFLSTM}). These approaches can be broadly categorized \cite{CLsurvey} into those leveraging the concepts of replay~\cite{iCArL, GEM, selectiveReplay}, regularization~\cite{ewc, regLwF,dynProject}, or parameter isolation~\cite{PackNet, progressiveNN}. But, all these approaches are infeasibly memory- and compute-intensive; e.g., they require separate per-task networks~\cite{hypernetworks}, computation of complex information metrics~\cite{ewc}, etc. 

In this paper, we show that using separate LSTM memory either for each task, or for each target label, can overcome catastrophic forgetting. State separation is effective because it implicitly achieves regularization and isolation of the hidden weights responsible for forgetting, without the explicit overheads of either. 
We validate our findings in two ways. First, we demonstrate our findings with computer memory access prefetching, an important sequential data processing problem used in computer system optimization. Compared to the state-of-the-art method to overcome catastrophic forgetting~\cite{ewc}, our approach achieves equal or better accuracy on sequential tasks with only half the training time. Second, we generalize our analysis to natural language processing, showing that state separation is effective across application domains. 

%% file: 2_Background.tex
\section{Background}
\label{back}

\textbf{LSTM Networks: }
LSTM networks~\cite{originalLSTM} are commonly used for learning on sequential data because they preserve information even across long sequences. A typical LSTM network takes in the current input symbol $\bm{x}_k$ and a state obtained from the previous step (consisting of the cell memory $\bm{C}_{k-1}$ and hidden state $\bm{h}_{k-1}$), to produce the current output symbol $\hat{\bm{y}_k}$ and the state that will be used in the next step. The LSTM update equations are shown below. The cell memory (and hidden state) are crucial for the sequential learning ability of the network.

\begin{align}
   \textrm{Forget gate, } \bm{f}_k &= \sigma(\bm{W_{fx}}.\bm{x}_k + \bm{W_{fh}}.\bm{h}_{k-1} + \bm{b_f}) 
   \label{eq:forget}\\
    \textrm{Input gate, } \bm{i}_k &= \sigma(\bm{W_{ix}}.\bm{x}_k + \bm{W_{ih}}.\bm{h}_{k-1} + \bm{b_i}) 
    \label{eq:input} \\
    \textrm{Output, } \hat{\bm{y}}_k  &= \sigma(\bm{W_{yx}}.\bm{x}_k + \bm{W_{yh}}.\bm{h}_{k-1} + \bm{b_y}) 
    \label{eq:output} \\
    \textrm{Cell memory, } \bm{C}_k &= \bm{f}_k*\bm{C}_{k-1} + \bm{i}_k*tanh(\bm{\bm{W_{Cx}}.\bm{x}_k + W_{Ch}}.\bm{h}_{k-1} + \bm{b_C}) \label{eq:cell}\\
      \textrm{Hidden state, } \bm{h}_k &=  \bm{y}_k*tanh(\bm{C}_k)
\end{align}

$(\bm{W_{ih}}, \bm{W_{fh}, \bm{W_{Ch}}, \bm{W_{yh}}})$ form the hidden weight matrix $\bm{W_{h}}$. $(\bm{W_{ii}}, \bm{W_{fi}, \bm{W_{Ci}}, \bm{W_{yi}}})$ forms the input weight matrix $\bm{W_{i}}$ \cite{pytorch}


In the continual learning setting, LSTM networks must learn a sequence of $n$ tasks, $\bm{T} = (T^j)_{j \in [1,n]}$. Each task $T^j = (\bm{X}^j, \bm{\Upsilon}^j)$ has a sequence of input symbols $\bm{X}^j = (\bm{x}^j_1, \bm{x}^j_2, \ldots, \bm{x}^j_{m_j})$  and a sequence of target output symbols $\bm{\Upsilon}^i = (\bm{y}^i_1, \bm{y}^i_2, \ldots, \bm{y}^i_{m_i})$ that the network must learn.

\textbf{Catastrophic Forgetting: } Catastrophic forgetting occurs because training a network on a new task overwrites the network's weights without preserving the information learned for older tasks. Catastrophic forgetting is not a problem when all the data and tasks that a network must learn are available offline, because the network can be trained to perform on all the tasks without forgetting in multiple epochs. But, many real-world scenarios require learning of new tasks on data previously un-encountered. Unfortunately, despite being first observed decades ago~\cite{frenchForgetting}, and studied extensively since~\cite{ewc,frenchForgetting, CLsurvey}, catastrophic forgetting remains an open problem. In fact, recent work has established that LSTM networks suffer from catastrophic forgetting just like other connectionist models~\cite{CFLSTM}. Ehret et al.~\cite{ehret2021continual} analyze RNNs to reveal that forgetting occurs because distinct tasks update the same shared weights shared by the LSTM state across timesteps. 



\textbf{The Memory Access Prefetching Problem: }
Memory accesses in a computer have orders of magnitude higher latency relative to the latency of a CPU's compute operation, resulting in the so-called 
Von Neumann\cite{vneumann} performance bottleneck for applications. This problem is becoming especially worse as the data processed by applications is increasing exponentially. Prefetching \cite{prefetching, hwprefetching} is an optimization that alleviates this bottleneck by predicting the next memory location accessed by an application, and bringing the data close to the CPU before it is actually requested~\cite{LearnMemAccPat}. 
Prefetching is possible because the memory accesses of an application have patterns such as a fixed stride access of consecutive elements in an array, arising from the application's algorithm or hardware memory layout. 

Recent works presented memory access prefetching as a sequence learning problem using LSTM networks that are trained offline\citep{LearnMemAccPat}. However, memory access patterns change over the course of an application, across applications, and with various input datasets to the applications. To be effective, a hardware or operating system prefetcher must learn these changing patterns online i.e., it must continually learn many patterns. Owing to its complexity and significance, we choose prefetching to demonstrate our work on mitigating catastrophic forgetting.

\textbf{Language Modelling:} Language modelling is a well-studied problem which involves tasks such as predicting the next word in a sentence. LSTMs have been used for natural language modelling \cite{LM_LSTM} due to their ability to maintaining temporal dependencies in sequences. Smaller LSTMs do not fare well on language modelling tasks. \cite{RNNreghw, RNNreg1} This necessites the need of larger LSTMs. However, larger LSTMs need techniques such as dropout \cite{RNNreghw, RNNreg1,OriginalDropout} to avoid overfitting. We choose to evaluate state separation as means to improve accuracy of smaller LSTMs by refining the information storage in the LSTM cell by state separation.

%% file: 3_RelatedWork.tex
\section{Related Work} \label{relwork} 
Catastrophic forgetting and its mitigation have been studied for many decades~\cite{CLsurvey,frenchForgetting,catastrophicInt}. Most of these focus on feed-forward networks, and can be grouped into  the following three broad categories~\citep{CLsurvey}. A detailed analysis is provided in~\citep{CLsurvey}.

\textbf{Replay-based}  (e.g.,~\citep{iCArL, GEM, selectiveReplay,Expreplay}):  These techniques interleave the training of a network with previous tasks when learning a new task, to prevent the old tasks from being forgotten. The samples used to train the old tasks could be the original ones stored in a replay memory~\cite{progressiveMem}, or produced using a generative network~\cite{selfRefresh}. 
Storing data scales poorly with increasing number of sequential tasks, while generative networks require significant compute. 

\textbf{Weight regularization} (e.g.,~\citep{ewc,regLwF,regEBLL}): Since catastrophic forgetting occurs because of changing the network weights without preserving old information, some approaches~\cite{ewc} target restricting the weight updates to prevent information loss. These approaches typically analyze the weights, e.g., with Fisher information~\citep{fisher}, to identify those that are critical to performance on an older task and prevent or cap their updates when learning on a new task. While this approach does not require storing past data, computing weight importance is resource-intensive. Furthermore, restricting updates to weights affects the plasticity of the network as the number of tasks increases, diminishing the network's learning ability\cite{ehret2021continual}. 

Alternative to computing task-related weight importance, some approaches (e.g.,~\cite{regLwF, rehearsal}) use knowledge distillation techniques to achieve regularization. While training on new tasks, these approaches use an additional term in the network loss function to ensure that the network reproduces previously learnt outputs for corresponding inputs, . 

\textbf{Parameter isolation} (e.g.,~\citep{PackNet, progressiveNN}): These approaches maintain different parameters like weights and entire sub-networks separate for each task, to completely avoid any interference in the weight updates. Optimizations like pruning~\cite{PackNet} and lateral connections between task-specific sub-networks~\cite{progressiveNN} reduce the number of network parameters that are used up, but compared to the other methods, this approach quickly saturates the network capacity for a given network size. 


There has been recent research establishing catastrophic forgetting in LSTM networks for sequential task learning~\cite{CFLSTM}, and studying the applicability of the above methods to alleviate it~\cite{ehret2021continual}. Duncker et al.~\cite{dynProject} use a weight regularization approach that allows large weight updates only when they corresponding input and output subspaces are orthogonal to those from previous tasks. Ehret et al.~\cite{ehret2021continual} have suggested that weight regularization~\cite{ewc}, and using task-specific hidden weights with hypernetworks\cite{hypernetworks} (i.e., parameter isolation) are promising to overcome catastrophic forgetting in LSTM networks.  Among these two approaches, they suggest that parameter isolation is superior because it does not limit the plasticity of the network. However, they have an overhead of outsourcing weight isolation to an external hypernetwork.

%% file: 4_Hypothesis.tex
\section{LSTM State Separation to Mitigate Catastrophic Forgetting}

The LSTM network's state retains the long term memory necessary to perform a task and is vital for its operation. We observe that reusing this memory across subsequent unrelated tasks diminishes performance on older ones. Further, we observe that the network's cell memory is a good filter for the network activations that are important for performance on a task. Therefore, we propose to separate the LSTM network's memory for each task protecting the network from catastrophic forgetting. For complex tasks with long sequences of symbols and correlations that are different from those in an earlier task, we propose state-separation based on each target label as an even better alternative. 

\begin{figure}[h]
\centering
\subfloat[A standard LSTM network.]{
\includegraphics[width=0.31\textwidth]{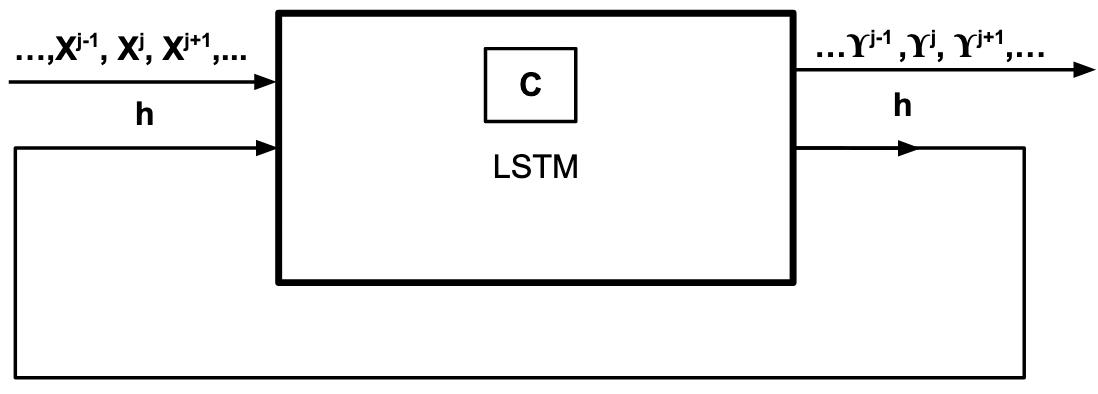}
\label{subfig_lstm}
}
\subfloat[State separation based on tasks.]{
\includegraphics[width=0.31\textwidth]{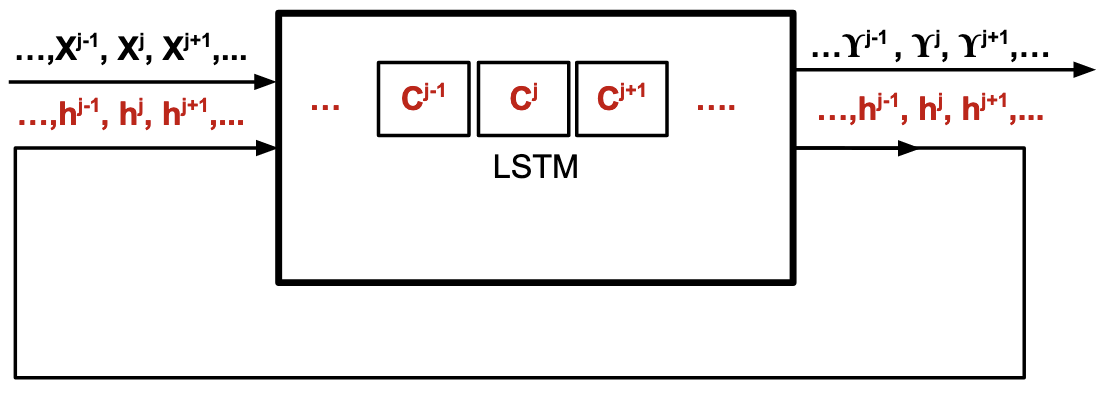}
\label{subfig_taskSep}
}%
\subfloat[State separation based on target label.]{
\includegraphics[width=0.31\textwidth]{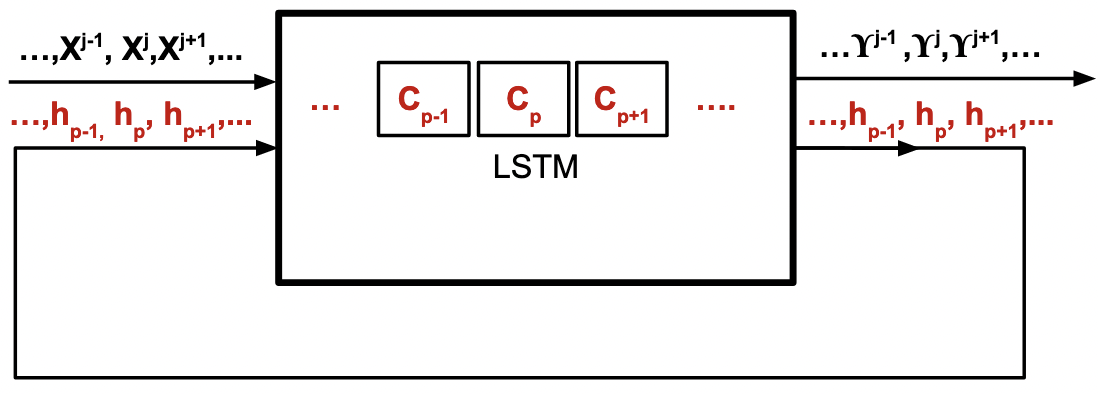}
\label{subfig_labelSep}
}
\caption{Overview of our proposed approaches to mitigate catastrophic forgetting, contrasted with a standard LSTM network.}
\label{proposed}
\end{figure}

Figure~\ref{proposed} illustrates our proposed methods. Figure~\ref{subfig_lstm} is a standard LSTM network where the same cell memory $\bm{C}$, and consequently the same hidden state $\bm{h}$, are reused across tasks. As a result, the LSTM is unable to retain performance on older tasks. With task-based state separation (Figure~\ref{subfig_taskSep}), we use a different cell memory $C^j$ and the hidden state $h^j$ for each unique task $T^j$. Since it is the cell memory that is key to the LSTM network's performance, separating it can overcome forgetting, even if the different tasks update the same shared set of weights (e.g., $\bm{W}_{Ch}$ from Equation~\ref{eq:cell}). Finally, Figure~\ref{subfig_labelSep} shows a finer-grained separation of the LSTM network's state based on the target label. Each hidden state $h_p$ corresponds to a unique target label. Intuitively, this approach seeks to protect the ability to recognize a particular label instead of the coarser-grained ability to perform a task, which may involve recognizing multiple labels.

Separating the states implicitly achieves weight regularization because samples from a new task use the task-specific state to update the weights more relevant to that task. It also implicitly achieves parameter isolation because there is a separate cell memory for each task. However, state separation does not require explicit calculations of task-related weight importance or subspace projections (e.g., as used in~\cite{ewc} or~\cite{dynProject}), or new hidden weights for each task (e.g., as suggested in~\cite{ehret2021continual, hypernetworks}). Separating the cell memory ($\bm{C}$) also requires an order of magnitude lesser storage than maintaining new weights ($\bm{W}_{Ch}$) per task.

%% file: 5_Experiment_setup.tex
\section{Experimental Setup}
\label{setup}


\textbf{Memory Access Prefetching: } We encode the address of each memory access as a difference (or $\delta$) from the previous. We define each task or a pattern to be a sequence of the $\delta$'s that an LSTM network should learn. Each task may have a different length of $\delta$'s, and there could be multiple occurrences of the same $\delta$ in a task or across tasks. For example, one task $T_1$ could be the sequence $(+2,+5,-7,+2)$ repeated again, while another task $T_2$ could be the sequence $(+2,+2,-3)$. The LSTM network prefetcher takes the current $\delta$ as input and predicts the next $\delta$ in the sequence. We construct synthetic sequences of various lengths and $\delta$'s to evaluate the prefetcher. We present the network with 10, 000 $\delta$'s in each task. 

Our prefetcher is a single layer LSTM with a hidden layer size of 100 and an input layer size of 100, implemented in PyTorch~\cite{pytorch}. We initialize the network weights with Xavier normalization\cite{pytorchLSTM}. We use the Adam optimizer~\cite{adam} with a learning rate of 0.001, and cross entropy loss given by  $H(x,y) = -\sum\limits_{x \in X} P(x) * log(P'(y))$, where 
$P$ is the probability that observation x occurs and $P'$ is the probability that target class y is predicted.


We perform two studies with this problem. The first is to show the ability of our approach to overcome forgetting. In this study, we select pairs of tasks and train the network on the first task for 100 epochs. Next, we introduce the second task for the network to learn, and monitor the accuracy of both tasks and the change in the network weights. We repeat the second task for several iterations to ensure that the network's performance on the second task is indeed stable as it is learnt.

In the second study, we evaluate the scalability of our method for increasing number of sequential tasks from 2 to 20. We also compare our method with Elastic Weight Consolidation (EWC)~\cite{ewc}, which is a state-of-the-art weight-regularization method to overcome catastrophic forgetting. We evaluate the networks using the total training time, and the mean accuracy over all tasks measured after the last task has been learnt~\cite{ehret2021continual}.
These was evaluations are performed on a server unit with 2 NVIDIA RTX 2080TI GPUs.

\textbf{Task Identification: } Our approach like prior work~\cite{ehret2021continual,ewc}, assumes that the arrival of a new task is known. Prior work has described many ways such as distribution tests, autoencoders, and explicit input, to obtain this information, and we do not consider the development of these methods in this paper.

\textbf{Language Models: } We use the word-level prediction task described in \citep{RNNreg1}, where the LSTM network must learn to predict the next word in a sequence of words. We use the Penn Tree Bank(PTB)\cite{ptb} dataset which has 10,000 words in its vocabulary. We use PTB's training set of 929k training words and 82k test words. 

We use a 2-layer LSTM with the size of hidden layer as 200 units and input layer as 200 units\cite{RNNreg1}. We train the network using the Adam Optimizer with a learning rate of 0.001. This evaluation is done on a cluster with 4 NVIDIA RTX 3090 GPUs. 

%% file: 6_Experimentation.tex
\section{Results}

\subsection{Continual Learning of Prefetching Task Pairs}

We consider three task pairs that respectively, have (i) the same  $\delta$'s in both tasks with same relative ordering but with different frequency of the occurence of $\delta$'s, (ii) different $\delta$'s in both tasks but with same  frequency of occurrence, and (iii) different $\delta$'s with different frequency of occurrence in both tasks. We devise these pairs to challenge  LSTM network with different types of similarity in the tasks, and observe if they lead to catastrophic forgetting.  

\begin{figure*}[t]
    \centering
    \begin{minipage}{.46\linewidth}
        \centering
        \subfloat[Same $\delta$, different frequency.]{
        \includegraphics[width=0.5\textwidth]{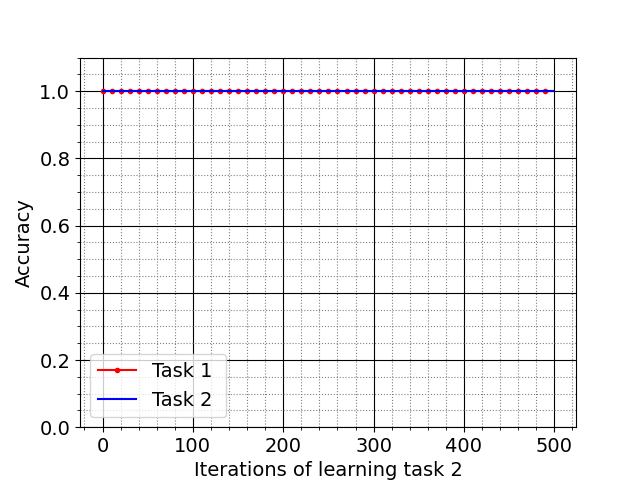}
        \label{1dfss_same_states_acc}
        }
        \subfloat[One extra $\delta$.]{
        \includegraphics[width=0.5\textwidth]{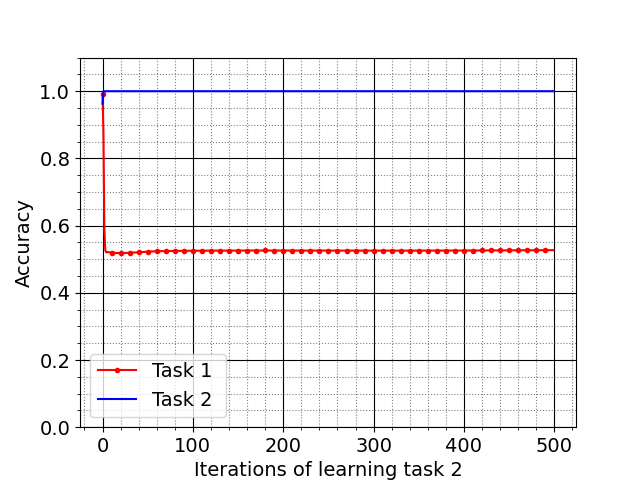}
        \label{3sym_dfss_same_state_acc}
        }
        \caption{Standard LSTM network.}
    \end{minipage} 
    \hspace{1mm}   
   \begin{minipage}{.46\linewidth}
        \centering
        \subfloat[Same $\delta$, different frequency.]{
        \includegraphics[width=0.5\textwidth]{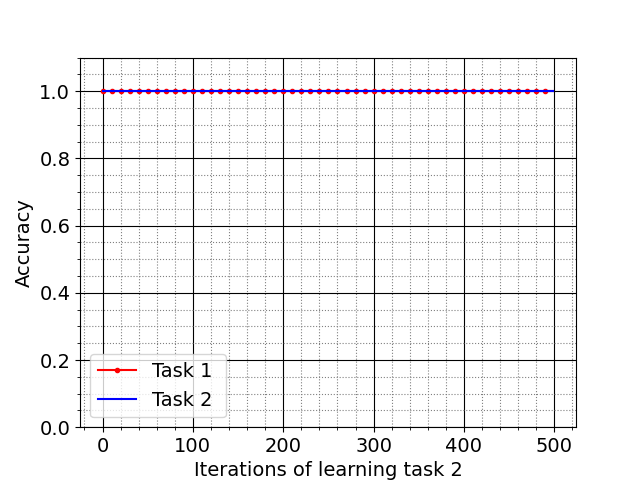}
        \label{1dfss_isolated_states_acc}
        }
        \subfloat[One extra $\delta$.]{
        \includegraphics[width=0.5\textwidth]{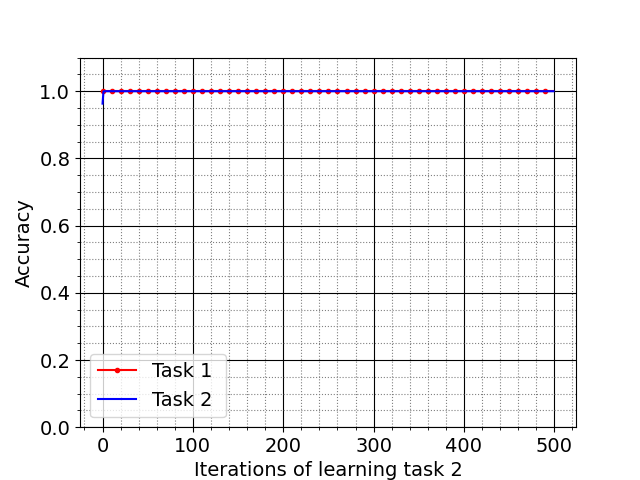}
        \label{3sym_dfss_sep_state_acc}
        }
        \caption{LSTM network with state separation.}
        \label{fig:sep_same_diff_1}
   \end{minipage}
\end{figure*}

\textbf{Same $\delta$, different frequency: } The pair of tasks is ($T_1$: +2, +2, -3, -3, -3, $\ldots$ | $T_2$: +2, +2, +2, +2, +2, -3, -3, -3, -3, -3, -3, -3, -3, $\ldots$). Both tasks have $\delta$ values of +2 and -3 but $T_1$ has 2 +2's followed by 3 -3's and $T_2$ has 5 +2's followed by 7 -3's. The LSTM network is pre-trained on $T_1$, and is exposed to $T_2$ online. Figure~\ref{1dfss_same_states_acc} shows the accuracy of the standard LSTM network on both tasks when the network learns an increasing number of $T_2$ iterations. Surprisingly, the standard LSTM network does not exhibit any forgetting in this scenario. This occurs because, with the same input and output symbols, the weight updates in the second task are in the same subspace of those of the first, allowing the network to retain its performance on the earlier task.
Catastrophic forgetting occurs only when the newer weight updates are in a different subspace than the earlier learnt ones. 

To confirm our insight, we run an additional experiment with the task pair ($T_1$: +2, +2, -3, -3, -3, $\ldots$ | $T_2$: +2, +2, -3, -3, -3, +7, $\ldots$) i.e., the two sequences differ by only one symbol. Figure~\ref{3sym_dfss_same_state_acc} shows that in this case, the performance on $T_1$ indeed falls abruptly within a few iterations. This confirms that even a slight change in the weight update subspaces can lead to catastrophic forgetting. However, the accuracy on $T_1$ does not go to 0 because there is overlap in the information between the two tasks.

Figure~\ref{1dfss_isolated_states_acc} and ~\ref{3sym_dfss_sep_state_acc} shows that using separate states per task has no catastrophic forgetting in both scenarios. Without separate states, the LSTM network gates eventually erase information the cell memory about the previous task since it is not relevant to the performance on the current task. Using a separate state for each task lets the network create a new memory for the current task, leaving the older cell memory unmodified. Consequently, the weight updates in the new task do not fully overlap with those for the previous, overcoming catastrophic forgetting.

\begin{figure*}[ht]
    \centering
    \begin{minipage}{.46\linewidth}
        \centering
        \subfloat[Different $\delta$, same frequency.]{
        \includegraphics[width=0.5\textwidth]{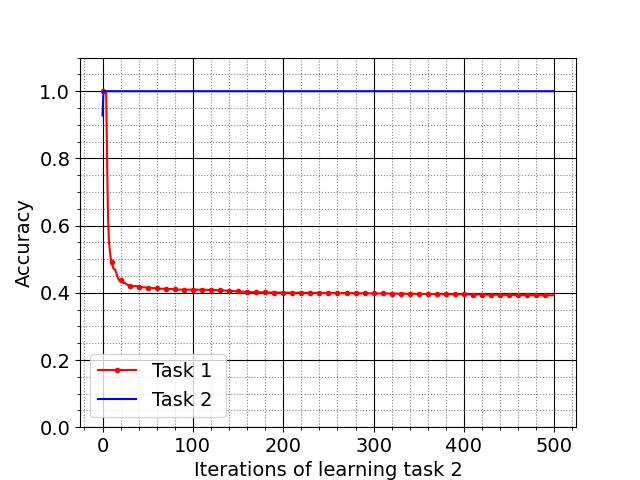}
        \label{sfds_same}
        }
        \subfloat[Different $\delta$, different frequency.]{
        \includegraphics[width=0.5\textwidth]{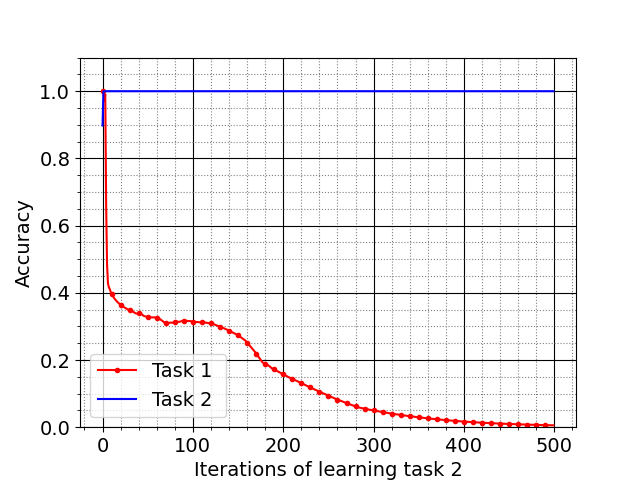}
        \label{dfds_same}
        }
        \caption{Standard LSTM network.}
    \end{minipage} \hspace{1mm}   
   \begin{minipage}{.46\linewidth}
        \centering
        \subfloat[Different $\delta$, same frequency.]{
        \includegraphics[width=0.5\textwidth]{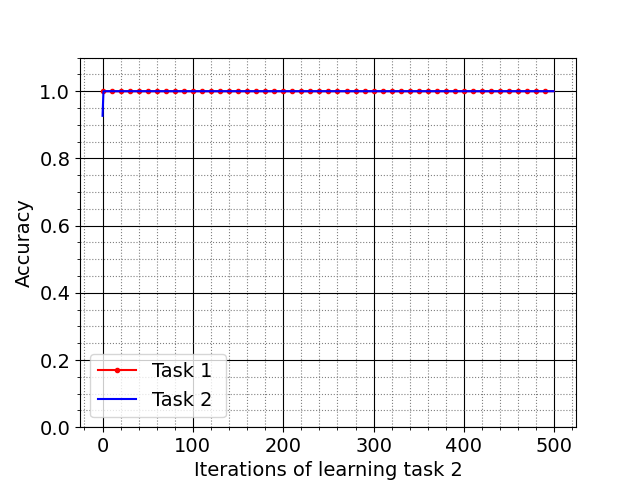}
        \label{sfds_isolated}
        }
        \subfloat[Different $\delta$, different frequency.]{
        \includegraphics[width=0.5\textwidth]{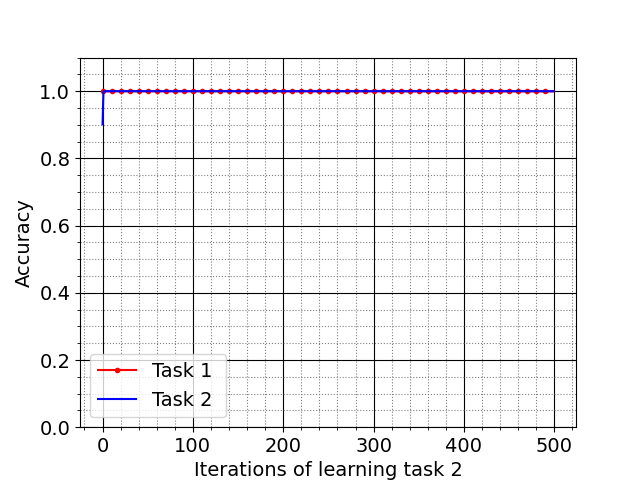}
        \label{dfds_isolated}
        }
        \caption{LSTM network with state separation.}
        \label{fig:sep_same_diff}
   \end{minipage}
\end{figure*}

\textbf{Different $\delta$, same frequency: }
We consider the task pair ($T_1$: +2, +2, -3, -3, -3, -3, -3, $\ldots$ | $T_2$: -7, -7, +5, +5, +5,$\ldots$), with different $\delta$'s in each task but their relative frequency of occurrence is the same i.e., $T_1$ it has 2 +2's followed by 3 -3's and $T_2$ has 2 -7's followed by 3 +5's. 
Figure~\ref{sfds_same} demonstrates forgetting in this case again. The forgetting is only partial because the two tasks have common information in terms of the relative frequency of the $\delta$'s. With separate states (Figure~\ref{sfds_isolated}), there is no forgetting as expected.

\textbf{Different $\delta$s, different frequency:} 
We consider task pair ($T_1$: +2, +2, +2, +2, +2, -3, -3, -3, -3, -3, -3, -3, -3, $\ldots$ | $T_2$: -7, -7, +5, +5, +5,$\ldots$), where there is no common feature between the tasks. Figure~\ref{dfds_same} shows that in addition to an abrupt fall in the accuracy for $T_1$, the network performance eventually goes to 0, as there is no shared information between the two tasks. Even in this challenging case, our approach (Figure~\ref{dfds_isolated}) overcomes catastrophic forgetting entirely, highlighting the effectiveness of our approach. 

\subsection{Scalability of State Separation and Comparison with EWC}
\begin{figure*}[t]
    \centering
    \begin{minipage}{.46\linewidth}
        \centering
        \subfloat[Accuracy with many sequential tasks.]{
        \includegraphics[width=0.5\textwidth]{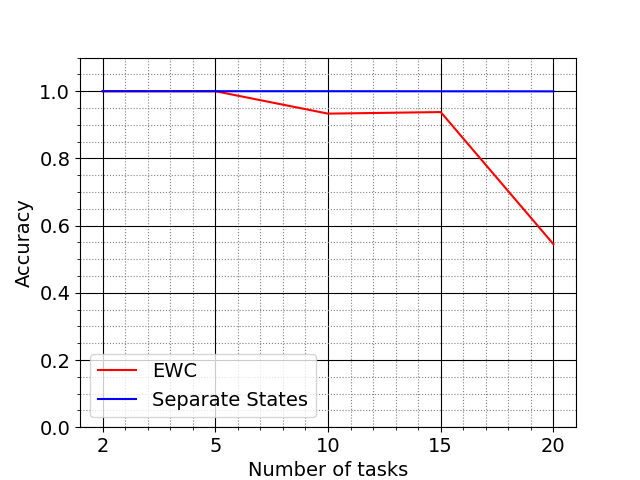}
        \label{ewc_acc}
        }
        \subfloat[Training time.]{
        \includegraphics[width=0.5\textwidth]{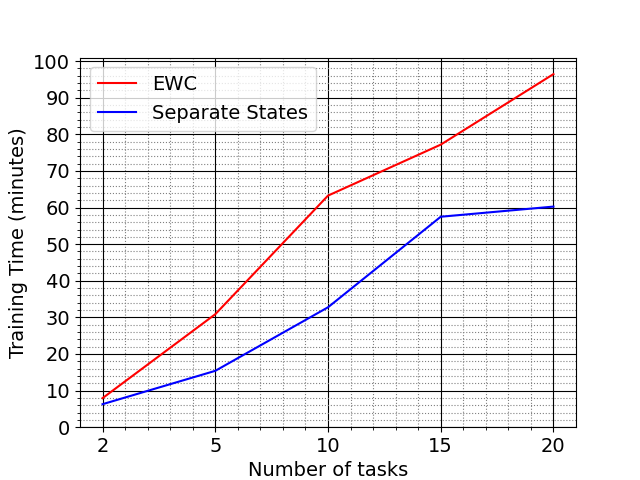}
        \label{ewc_time}
        }
        \caption{Comparing with EWC.}
        \label{fig:ewc}
    \end{minipage} \hspace{1mm}   
   \begin{minipage}{.46\linewidth}
        \centering
        \subfloat[Standard LSTM network.]{
        \includegraphics[width=0.5\textwidth]{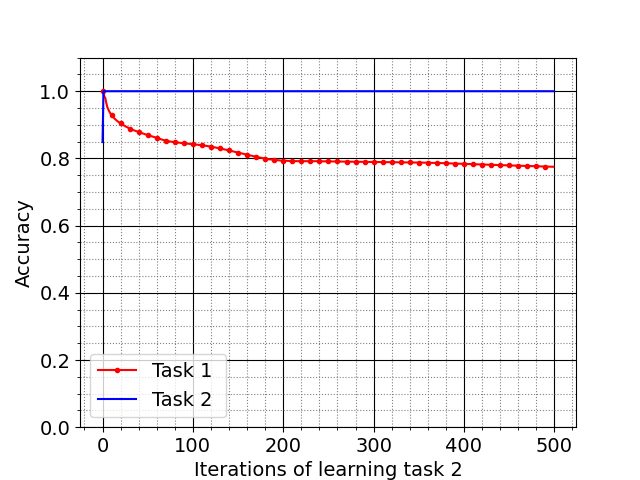}
        \label{weight_same}
        }
        \subfloat[LSTM network with state separation.]{
        \includegraphics[width=0.5\textwidth]{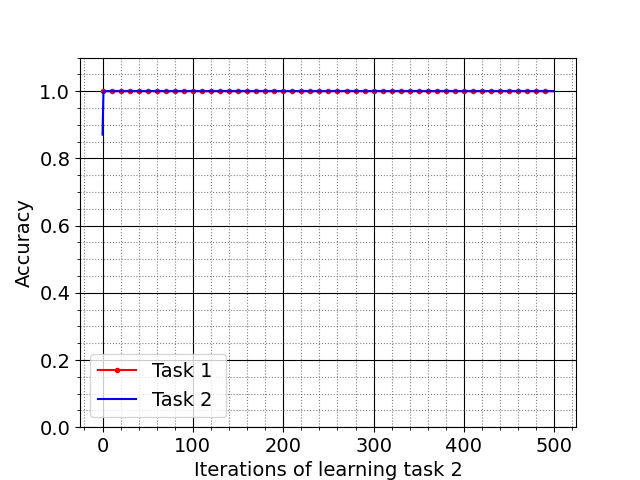}
        \label{weight_isolated}
        }
        \caption{Impact of Uniform Weight Initialization.}
        \label{fig:weight}
   \end{minipage}
\end{figure*}

We study how our proposal performs for an increasing number of sequential tasks, and compare it with that of EWC, a state-of-the-art method to mitigate catastrophic forgetting. Similar to our approach EWC only needs the task identifier to compute performance-preserving weight updates. 
We test our approach and EWC with 2--20 sequential tasks, each with a single $\delta$ repeated 10, 000 times as before. Figure~\ref{fig:ewc} shows the mean task accuracy and training time of both approaches in this study. The mean accuracy of EWC begins to fall around 6 tasks because the network  weights lose their plasticity to learn new tasks\cite{ehret2021continual}. EWC prevents newer tasks from changing the network weights that are deemed important for earlier tasks.  As a result, when more tasks are introduced, there is less flexibility in the network to learn them. As a result, network performance suffers. In contrast, state separation does not suffer from this limitation since it does not restrict weight updates. Hence, it can preserve performance for a higher number of tasks. 

Figure~\ref{ewc_time} shows that EWC takes about twice as long as our approach to train since it must calculate weight importance whenever a new task arrives, which is computationally expensive~\cite{fisher}. Worse, this overhead grows as more tasks are presented to the system. On the other hand, state separation by itself does not require any computation, and scales better.

\subsection{Role of Weight Initialization}

Since separating cell memory implicitly influences updates to the weights, we study if weight initialization has a role in the effectiveness of our approach, and on catastrophic forgetting broadly. Apart from the default  Xavier initialization, we initialize the network with values from a uniform random distribution over $[-1,1]$. This range of values is wider than those available in the default method, and may allow the network to converge with fewer weight updates. We test the network with the different $\delta$, different frequencies task pair described earlier.

Figure~\ref{weight_same} shows that uniform initialization results in forgetting, but the degree is significantly improved over the default Xavier initialization---compare with Figure~\ref{dfds_same}. We attribute this improvement to the fewer weight updates that are made with the new initialization method. Nonetheless, changing initialization does not prevent forgetting and the accuracy continues to deteriorate, although only slowly. However, we note that the new weight initialization does not diminish the performance of state separation (Figure~\ref{weight_isolated}), since our approach works by protecting from the erasure of cell memory.

    
\subsection{Complex Patterns and Granularity of State Separation}

\begin{figure}[t]
\centering
\subfloat[Standard LSTM Network.]{
\includegraphics[width=0.3\textwidth]{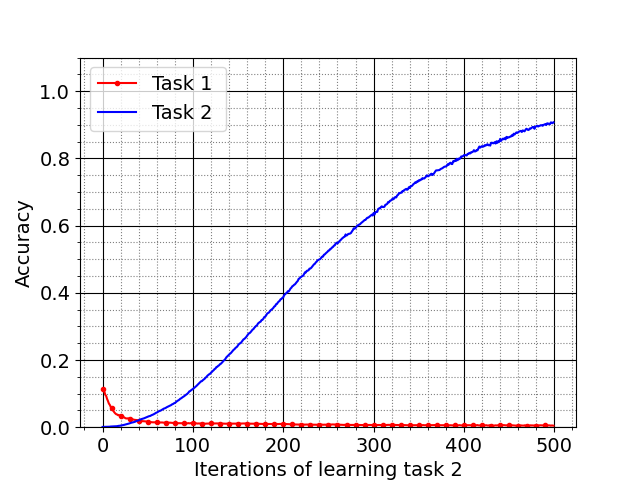}
\label{ptr_same_state_acc}
}%
\subfloat[Task-based state separation.]{
\includegraphics[width=0.3\textwidth]{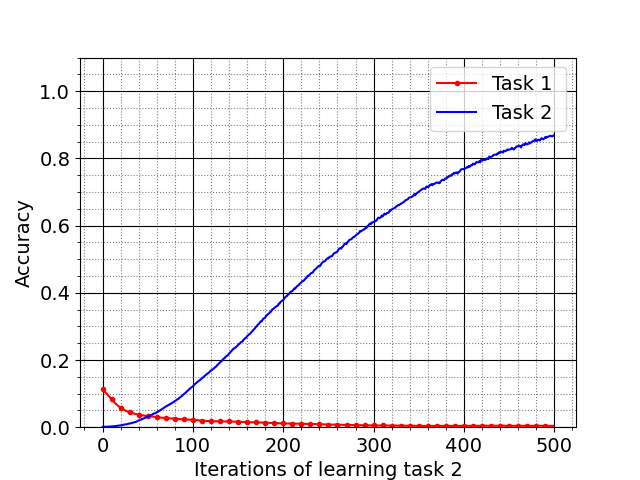}
\label{ptr_isolated_states_per_task_acc}
}
\subfloat[Label-based state separation.]{
\includegraphics[width=0.3\textwidth]{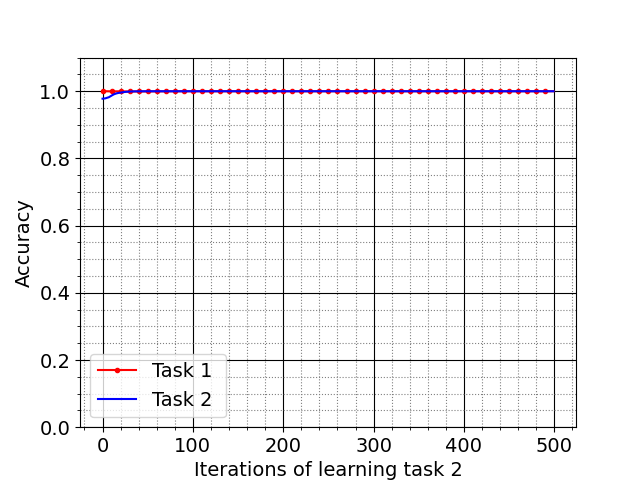}
\label{ptr_isolated_states_acc}
}
\caption{Studying the efficacy of standard and proposed LSTM techniques for complex tasks.}
\label{fig_overview}
\end{figure}

We create a task pair where the $\delta$'s in each task are drawn randomly from a set of 1, 000 pseudorandom integers. Such irregular sequences occur frequently in the memory accesses of pointer-based applications. Figure~\ref{ptr_same_state_acc} shows that the standard LSTM network experiences catastrophic forgetting as expected. Given the complexity of the task, the network accuracy falls off with just one iteration of the second task.

Figure~\ref{ptr_isolated_states_per_task_acc} shows that task-based state separation also experiences catastrophic forgetting. This exposes the weakness in task-based separation that by separating cell memories between tasks, the method only implicitly separates the network weights. This approach works when there is a non-random underlying distribution in the tasks such that the network's memory in each task has a different affinity with the weights. When the tasks have pseudorandom inputs, the cell memory is a weak proxy to separate weight interference, and task-based state-separation is ineffective. 

In these cases, we propose using a finer-grained separation of states based on the output target labels instead of tasks. Even when each task is composed of arbitrary complex inputs, as long as there is a distribution that generates outputs from them, separating the network memory based on the output labels will be effective in protecting the weights important for the corresponding labels. Indeed, Figure~\ref{ptr_isolated_states_acc} shows that label-based state separation does not suffer from catastrophic forgetting for the task pair with pseudorandom $\delta$ values.

Overall, we find that state separation is an effective technique to mitigate catastrophic forgetting in LSTMs. State separation works by implicitly protecting the network weights from losing information learnt from previous tasks. When tasks are complex with arbitrary input symbols, finer-grained state separation based on the output labels is effective. State separation is faster and requires less memory than alternative methods, which require explicit weight-importance calculations, or additional networks. 

\subsection{An Additional Use of State Separation: Offline Language Modelling}

We evaluate the use of state separation mechanism in LSTMs for smaller non-regularized LSTMs in language modelling tasks. We measure the network's ability to predict the correct next word in the test set after it was trained using PTB's training set. Note that the LSTM we evaluate is small and is typically deemed insufficient for performance on the language modelling task\cite{RNNreg1}. 

Our technique using state separation per target label achieved an accuracy of 0.9 compared to the standard LSTM which achieved an accuracy of 0.05.  This is because the separated states maintain memory for each target label. In the standard LSTM scenario, the same cell memory gets updated with a lot of information which diminishes the importance of prior information leading to forgetting. The proposed technique can enable use of smaller LSTMs for language modelling without the need of using larger LSTMs and regularization schemes like dropout\cite{RNNreg1} to avoid overfitting.

%% file: Conclusion.tex
\section{Conclusion}
We present a novel strategy for reducing catastrophic forgetting in LSTMs by using the underlying LSTM mechanisms. This approach does not necessitate the building or outsourcing of any external network for regularization, nor does it necessitate the implementation of any explicit regularization scheme. The solution is simple, effective, and allows for faster learning.

%% file: 7_SupplementaryMaterial.tex
\section*{Supplementary Material}

\section{Continual Learning of Prefetching Task Pairs}

Section 6.1 explains how a minor change in the weight update subspace causes forgetting. Consider a task pair ($T_1$: +2, +2, -3, -3, -3, $\ldots$ | $T_2$: +2, +2, -3, -3, -3, +7, $\ldots$) i.e., the two sequences differ by only one $\delta$ symbol i.e. one +7 is considered. Figure~\ref{3sym_dfss_same_state_acc} demonstrates how forgetting happens in a standard LSTM even when a single extra $\delta$ symbol i.e. +7 is added to otherwise identical patterns. Figure~\ref{3sym_dfss_sep_state_acc} shows separating states in LSTM helps maintain accuracy even in this case.

\subsection{Same \texorpdfstring{$\delta$}, different frequency}
\begin{figure*}[!htbp] 
    \begin{minipage}{.46\linewidth}
        \centering
        \subfloat[Two extra $\delta$s.]{
        \includegraphics[width=0.5\textwidth]{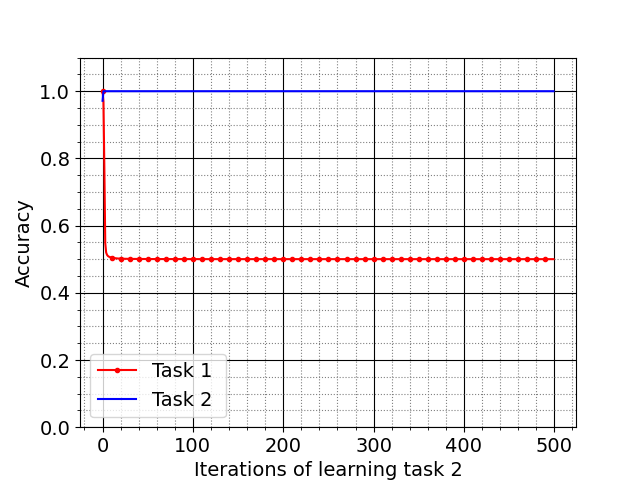}
        \renewcommand{\thefigure}{\arabic{figure}}
        \label{3symbols_2extra_std}
        }
        \subfloat[Three extra $\delta$s.]{
        \includegraphics[width=0.5\textwidth]{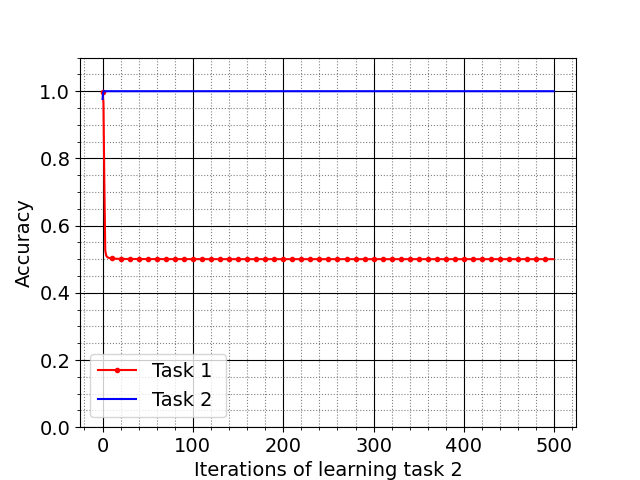}     \label{3symbols_3extra_std}
        }
        \subfloat[Four extra $\delta$s.]{
        \includegraphics[width=0.5\textwidth]{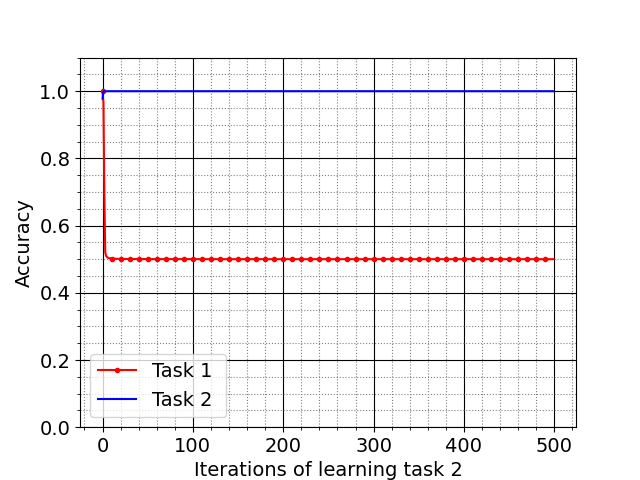}     \label{3symbols_4extra_std}
        }
        \subfloat[Five extra $\delta$s.]{
        \includegraphics[width=0.5\textwidth]{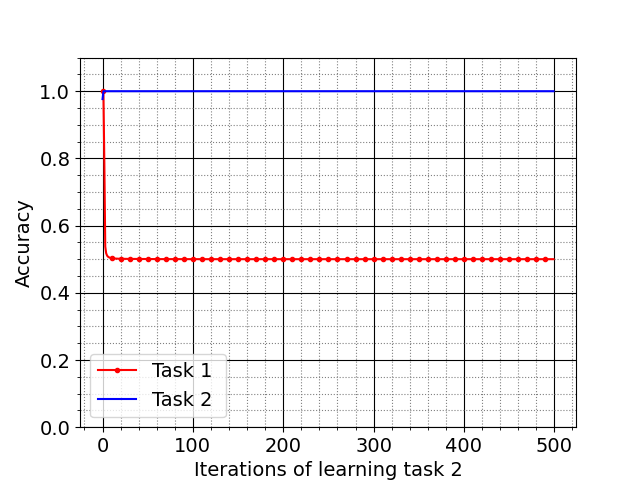}     \label{3symbols_5extra_std}
        }
    \end{minipage}
    \caption{Standard LSTM network.}
    \label{3symbols_std}
\end{figure*}
\begin{figure*}[!htbp] 
    \begin{minipage}{.46\linewidth}
        \centering
        \subfloat[Two extra $\delta$s.]{
        \includegraphics[width=0.5\textwidth]{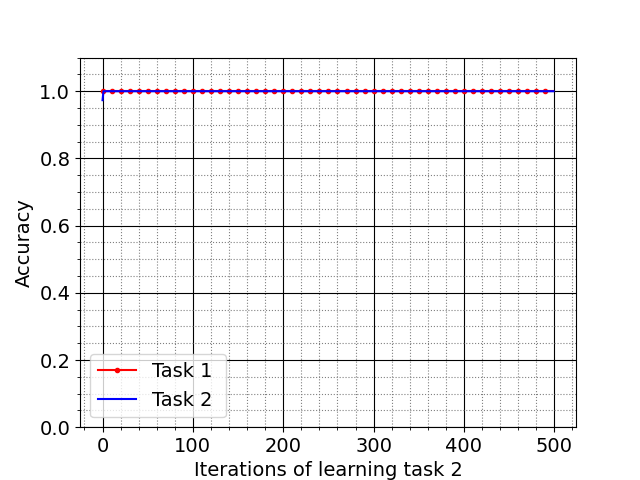}
        \label{3symbols_2extra_sepState}
        }
        \subfloat[Three extra $\delta$s.]{
        \includegraphics[width=0.5\textwidth]{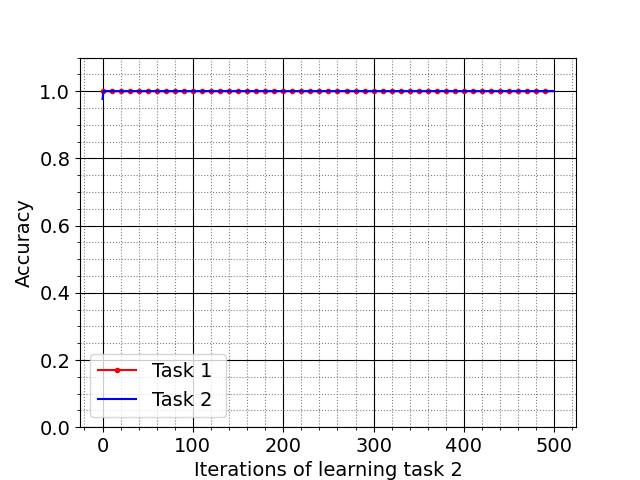}     \label{3symbols_3extra_sepState}
        }
        \subfloat[Four extra $\delta$s.]{
        \includegraphics[width=0.5\textwidth]{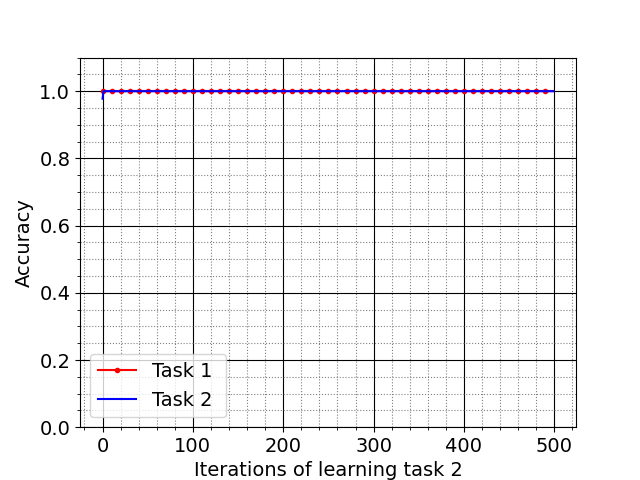}     \label{3symbols_4extra_sepState}
        }
        \subfloat[Five extra $\delta$s.]{
        \includegraphics[width=0.5\textwidth]{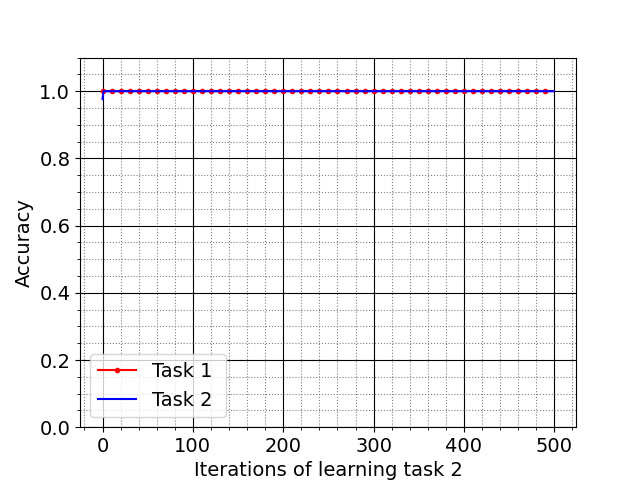}     \label{3symbols_5extra_sepState}
        }
    \end{minipage}
    \caption{LSTM network with state separation.}
    \label{3symbols_extra_sepState}
    
\end{figure*}

We extend this experiment to see if forgetting rises as the difference between the patterns increases in terms of the frequency of occurrence of the different $\delta$ symbol, i.e. +7. We increase the number of +7 $\delta$s from 2 to 5. Figure~\ref{3symbols_std} demonstrates the same abrupt drop in performance on $T1$ regardless of the frequency of occurrence of the differing $\delta$ symbol. This means that little changes in the weight update subspace induced by minor changes in the input are equivalent to larger instances of the change in the input causing a shift in the weight update subspace, which results in forgetting. Figure~\ref{3symbols_extra_sepState} demonstrates that separating states preserves prior information in all scenarios of increasing the occurrence of a different symbol. Using different states allows the network to build a new memory for the new task while leaving the older cell memory unchanged. The new task's weight updates do not overlap the previous updates, preventing catastrophic forgetfulness.

\section{Online Learning in Language Models}
We evaluate the benefit of separating states in LSTMs for online learning in natural language applications. We select a word prediction task\cite{RNNreg1}. Retaining context of immediate past information such as the immediately preceding few sentences such as the last paragraph in natural language applications is crucial to retaining context across the generated text. Each paragraph is treated as a task. To assess the capacity of the regular LSTM and state separated LSTM to maintain context, i.e. to recall the previous paragraph, we measure the network's ability to recall the first paragraph (task $T1$) after training on the next paragraph (task $T2$). The network is not pre-trained on the vocabulary. The succeeding paragraphs are selected form PTB's\cite{ptb} training set. With LSTM network with separate states, at the end of 20 epochs, the second task $T2$ achieves an accuracy of 0.87 while it is able to recall the preceding paragraph, $T1$,  with an accuracy of 0.76. While the standard LSTM achieves an accuracy of 0.72 on the new paragraph, $T2$ and recalls   $T1$ with an accuracy of 0.06 on the end of 20 epochs. This demonstrates that separating the cell memory per target label allows maintaining information per word for next word prediction. Cell memory is segregated per label in the LSTM with different states, which protects the weights important for the corresponding label from being updated on learning new information. In a standard LSTM scenario, the cell memory is updated with new information, which diminishes the relevance of previous information, causing forgetting.


%% file: main.bbl
\begin{thebibliography}{10}

\bibitem{medicalLSTM}
Eric~J Topol.
\newblock High-performance medicine: the convergence of human and artificial
  intelligence.
\newblock {\em Nature medicine}, 25(1):44--56, 2019.

\bibitem{weatherLSTM}
Zahra Karevan and Johan~A.K. Suykens.
\newblock Transductive lstm for time-series prediction: An application to
  weather forecasting.
\newblock {\em Neural Networks}, 125:1--9, 2020.

\bibitem{stockMarketLSTMPrediction}
Jiaqi Li, Xiaoyan Wang, Saleem Ahmad, Xiaobing Huang, and Yousaf~Ali Khan.
\newblock Optimization of investment strategies through machine learning.
\newblock {\em Heliyon}, page e16155, 2023.

\bibitem{languageModellingWithLSTM}
Martin Sundermeyer, Hermann Ney, and Ralf Schlüter.
\newblock From feedforward to recurrent lstm neural networks for language
  modeling.
\newblock {\em IEEE/ACM Transactions on Audio, Speech, and Language
  Processing}, 23(3):517--529, 2015.

\bibitem{LearnMemAccPat}
Milad Hashemi, Kevin Swersky, Jamie Smith, Grant Ayers, Heiner Litz, Jichuan
  Chang, Christos Kozyrakis, and Parthasarathy Ranganathan.
\newblock Learning memory access patterns.
\newblock In Jennifer Dy and Andreas Krause, editors, {\em Proceedings of the
  35th International Conference on Machine Learning}, volume~80 of {\em
  Proceedings of Machine Learning Research}, pages 1919--1928. PMLR, 10--15 Jul
  2018.

\bibitem{mlPrefetchervoyager}
Zhan Shi, Akanksha Jain, Kevin Swersky, Milad Hashemi, Parthasarathy
  Ranganathan, and Calvin Lin.
\newblock A hierarchical neural model of data prefetching.
\newblock In {\em Proceedings of the 26th ACM International Conference on
  Architectural Support for Programming Languages and Operating Systems},
  ASPLOS '21, page 861–873, New York, NY, USA, 2021. Association for
  Computing Machinery.

\bibitem{ehret2021continual}
Benjamin Ehret, Christian Henning, Maria Cervera, Alexander Meulemans,
  Johannes~Von Oswald, and Benjamin~F Grewe.
\newblock Continual learning in recurrent neural networks.
\newblock In {\em International Conference on Learning Representations}, 2021.

\bibitem{CFLSTM}
Monika Schak and Alexander Gepperth.
\newblock A study on catastrophic forgetting in deep lstm networks.
\newblock In Igor~V. Tetko, V{\v{e}}ra K{\r{u}}rkov{\'a}, Pavel Karpov, and
  Fabian Theis, editors, {\em Artificial Neural Networks and Machine Learning
  -- ICANN 2019: Deep Learning}, pages 714--728, Cham, 2019. Springer
  International Publishing.

\bibitem{iCArL}
S.~Rebuffi, A.~Kolesnikov, G.~Sperl, and C.~H. Lampert.
\newblock icarl: Incremental classifier and representation learning.
\newblock In {\em 2017 IEEE Conference on Computer Vision and Pattern
  Recognition (CVPR)}, pages 5533--5542, Los Alamitos, CA, USA, jul 2017. IEEE
  Computer Society.

\bibitem{ewc}
James Kirkpatrick, Razvan Pascanu, Neil Rabinowitz, Joel Veness, Guillaume
  Desjardins, Andrei~A Rusu, Kieran Milan, John Quan, Tiago Ramalho, Agnieszka
  Grabska-Barwinska, et~al.
\newblock Overcoming catastrophic forgetting in neural networks.
\newblock {\em Proceedings of the national academy of sciences},
  114(13):3521--3526, 2017.

\bibitem{PackNet}
Arun Mallya and Svetlana Lazebnik.
\newblock Packnet: Adding multiple tasks to a single network by iterative
  pruning.
\newblock {\em 2018 IEEE/CVF Conference on Computer Vision and Pattern
  Recognition}, pages 7765--7773, 2017.

\bibitem{CLsurvey}
Matthias De~Lange, Rahaf Aljundi, Marc Masana, Sarah Parisot, Xu~Jia, Aleš
  Leonardis, Gregory Slabaugh, and Tinne Tuytelaars.
\newblock A continual learning survey: Defying forgetting in classification
  tasks.
\newblock {\em IEEE Transactions on Pattern Analysis and Machine Intelligence},
  44(7):3366--3385, 2022.

\bibitem{GEM}
David Lopez-Paz and Marc'Aurelio Ranzato.
\newblock Gradient episodic memory for continual learning.
\newblock In {\em Proceedings of the 31st International Conference on Neural
  Information Processing Systems}, NIPS'17, page 6470–6479, Red Hook, NY,
  USA, 2017. Curran Associates Inc.

\bibitem{selectiveReplay}
David Isele and Akansel Cosgun.
\newblock Selective experience replay for lifelong learning.
\newblock In {\em Proceedings of the Thirty-Second AAAI Conference on
  Artificial Intelligence and Thirtieth Innovative Applications of Artificial
  Intelligence Conference and Eighth AAAI Symposium on Educational Advances in
  Artificial Intelligence}, AAAI'18/IAAI'18/EAAI'18. AAAI Press, 2018.

\bibitem{regLwF}
Zhizhong Li and Derek Hoiem.
\newblock Learning without forgetting.
\newblock In Bastian Leibe, Jiri Matas, Nicu Sebe, and Max Welling, editors,
  {\em Computer Vision -- ECCV 2016}, pages 614--629, Cham, 2016. Springer
  International Publishing.

\bibitem{dynProject}
Lea Duncker, Laura Driscoll, Krishna~V Shenoy, Maneesh Sahani, and David
  Sussillo.
\newblock Organizing recurrent network dynamics by task-computation to enable
  continual learning.
\newblock In H.~Larochelle, M.~Ranzato, R.~Hadsell, M.F. Balcan, and H.~Lin,
  editors, {\em Advances in Neural Information Processing Systems}, volume~33,
  pages 14387--14397. Curran Associates, Inc., 2020.

\bibitem{progressiveNN}
Andrei~A. Rusu, Neil~C. Rabinowitz, Guillaume Desjardins, Hubert Soyer, James
  Kirkpatrick, Koray Kavukcuoglu, Razvan Pascanu, and Raia Hadsell.
\newblock Progressive neural networks, 2022.

\bibitem{hypernetworks}
David Ha, Andrew~M. Dai, and Quoc~V. Le.
\newblock Hypernetworks.
\newblock In {\em International Conference on Learning Representations}, 2017.

\bibitem{originalLSTM}
Sepp Hochreiter and Jürgen Schmidhuber.
\newblock Long short-term memory.
\newblock {\em Neural Computation}, 9(8):1735--1780, 1997.

\bibitem{pytorch}
{Pytorch}, howpublished = {\url{https://pytorch.org/}}, note = {Accessed:
  2023-17-05}.

\bibitem{frenchForgetting}
Robert~M. French.
\newblock Catastrophic forgetting in connectionist networks.
\newblock {\em Trends in Cognitive Sciences}, 3(4):128--135, 1999.

\bibitem{vneumann}
II~Arikpo, FU~Ogban, and IE~Eteng.
\newblock Von neumann architecture and modern computers.
\newblock {\em Global Journal of Mathematical Sciences}, 6(2):97--103, 2007.

\bibitem{prefetching}
David Callahan, Ken Kennedy, and Allan Porterfield.
\newblock Software prefetching.
\newblock In {\em Proceedings of the Fourth International Conference on
  Architectural Support for Programming Languages and Operating Systems},
  ASPLOS IV, page 40–52, New York, NY, USA, 1991. Association for Computing
  Machinery.

\bibitem{hwprefetching}
Babak Falsafi and Thomas~F. Wenisch.
\newblock A primer on hardware prefetching.
\newblock In {\em A Primer on Hardware Prefetching}, 2014.

\bibitem{LM_LSTM}
Daniel Soutner and Lud{\v{e}}k M{\"u}ller.
\newblock Application of lstm neural networks in language modelling.
\newblock In Ivan Habernal and V{\'a}clav Matou{\v{s}}ek, editors, {\em Text,
  Speech, and Dialogue}, pages 105--112, Berlin, Heidelberg, 2013. Springer
  Berlin Heidelberg.

\bibitem{RNNreghw}
Vu~Pham, Théodore Bluche, Christopher Kermorvant, and Jérôme Louradour.
\newblock Dropout improves recurrent neural networks for handwriting
  recognition.
\newblock In {\em 2014 14th International Conference on Frontiers in
  Handwriting Recognition}, pages 285--290, 2014.

\bibitem{RNNreg1}
Wojciech Zaremba, Ilya Sutskever, and Oriol Vinyals.
\newblock Recurrent neural network regularization, 2014.

\bibitem{OriginalDropout}
Nitish Srivastava.
\newblock Improving neural networks with dropout.
\newblock {\em University of Toronto}, 182(566):7, 2013.

\bibitem{catastrophicInt}
Michael McCloskey and Neal~J. Cohen.
\newblock Catastrophic interference in connectionist networks: The sequential
  learning problem.
\newblock {\em Psychology of Learning and Motivation}, 24:109--165, 1989.

\bibitem{Expreplay}
David Rolnick, Arun Ahuja, Jonathan Schwarz, Timothy Lillicrap, and Gregory
  Wayne.
\newblock Experience replay for continual learning.
\newblock In H.~Wallach, H.~Larochelle, A.~Beygelzimer, F.~d\textquotesingle
  Alch\'{e}-Buc, E.~Fox, and R.~Garnett, editors, {\em Advances in Neural
  Information Processing Systems}, volume~32. Curran Associates, Inc., 2019.

\bibitem{progressiveMem}
Nabiha Asghar, Lili Mou, Kira~A Selby, Kevin~D Pantasdo, Pascal Poupart, and
  Xin Jiang.
\newblock Progressive memory banks for incremental domain adaptation.
\newblock {\em arXiv preprint arXiv:1811.00239}, 2018.

\bibitem{selfRefresh}
Bernard Ans, St{\'e}phane Rousset, Robert~M French, and Serban Musca.
\newblock Self-refreshing memory in artificial neural networks: Learning
  temporal sequences without catastrophic forgetting.
\newblock {\em Connection Science}, 16(2):71--99, 2004.

\bibitem{regEBLL}
Daniel~L. Silver and Robert~E. Mercer.
\newblock The task rehearsal method of life-long learning: Overcoming
  impoverished data.
\newblock In Robin Cohen and Bruce Spencer, editors, {\em Advances in
  Artificial Intelligence}, pages 90--101, Berlin, Heidelberg, 2002. Springer
  Berlin Heidelberg.

\bibitem{fisher}
Alexander Soen and Ke~Sun.
\newblock On the variance of the fisher information for deep learning.
\newblock In M.~Ranzato, A.~Beygelzimer, Y.~Dauphin, P.S. Liang, and J.~Wortman
  Vaughan, editors, {\em Advances in Neural Information Processing Systems},
  volume~34, pages 5708--5719. Curran Associates, Inc., 2021.

\bibitem{rehearsal}
Daniel~L. Silver and Robert~E. Mercer.
\newblock The task rehearsal method of life-long learning: Overcoming
  impoverished data.
\newblock In Robin Cohen and Bruce Spencer, editors, {\em Advances in
  Artificial Intelligence}, pages 90--101, Berlin, Heidelberg, 2002. Springer
  Berlin Heidelberg.

\bibitem{pytorchLSTM}
{Pytorch LSTM module}, howpublished =
  {\url{https://pytorch.org/docs/stable/generated/torch.nn.lstm.html}}, note =
  {Accessed: 2023-17-05}.

\bibitem{adam}
Diederik~P. Kingma and Jimmy Ba.
\newblock Adam: {A} method for stochastic optimization.
\newblock In Yoshua Bengio and Yann LeCun, editors, {\em 3rd International
  Conference on Learning Representations, {ICLR} 2015, San Diego, CA, USA, May
  7-9, 2015, Conference Track Proceedings}, 2015.

\bibitem{ptb}
Mitchell~P. Marcus, Beatrice Santorini, and Mary~Ann Marcinkiewicz.
\newblock Building a large annotated corpus of {E}nglish: The {P}enn
  {T}reebank.
\newblock {\em Computational Linguistics}, 19(2):313--330, 1993.

\end{thebibliography}
